\documentclass[10pt,twocolumn,letterpaper]{article}
\usepackage{cvpr}
\usepackage{times}

\usepackage{cite}
\usepackage{url}
\usepackage{graphicx}
\usepackage{amsmath}
\usepackage{amssymb}
\usepackage{pifont} % For checkmark and xmark
\usepackage{bm} %to have superscript at the right position when using \boldsymbol. Must be loaded after amsbsy, which is loaded by amsmath
\usepackage{subfig}
\usepackage[colorlinks=true,bookmarks=false]{hyperref} %hyperref should go last

\usepackage[noabbrev,nameinlink,capitalise]{cleveref} %... but cleveref should go after hyperref

\graphicspath{ {./images/} }

\newcommand{\cmark}{\ding{51}~}%
\newcommand{\xmark}{\ding{55}~}%

\cvprfinalcopy

\title{Zero-Shot Human-Object Interaction Recognition via Affordance Graphs}

\begin{document}
	
%	\IEEEoverridecommandlockouts
%	\IEEEpubid{\makebox[\columnwidth]{%INSERT DOI HERE
%			~\copyright{}2019 IEEE \hfill} \hspace{\columnsep}\makebox[\columnwidth]{ }
%		}
	
	\title{Zero-Shot Human-Object Interaction Recognition via Affordance Graphs}
	
	\author{
		Alessio Sarullo\\
		\textit{Department of Computer Science} \\
		\textit{University of Manchester}\\
		Manchester, UK \\
		alessio.sarullo@manchester.ac.uk
		\and
		Tingting Mu\\
		\textit{Department of Computer Science} \\
		\textit{University of Manchester}\\
		Manchester, UK \\
		tingting.mu@manchester.ac.uk
	}

\maketitle

\begin{abstract}
	We propose a new approach for Zero-Shot Human-Object Interaction Recognition in the challenging setting that involves interactions with unseen actions (as opposed to just unseen combinations of seen actions and objects). Our approach makes use of knowledge external to the image content in the form of a graph that models affordance relations between actions and objects, i.e., whether an action can be performed on the given object or not. We propose a loss function with the aim of distilling the knowledge contained in the graph into the model, while also using the graph to regularise learnt representations by imposing a local structure on the latent space. We evaluate our approach on several datasets (including the popular HICO and HICO-DET) and show that it outperforms the current state of the art.
\end{abstract}

\section{Introduction}
Human-Object Interaction (HOI) Recognition is the task of identifying how people interact with the surrounding objects from the visual appearance of the scene and it is of paramount importance to understand the content of an image. It consists of producing a set of $ \langle human, action, object \rangle $ triplets for the input image, providing a concise representation of the image semantics that can be used in higher-level tasks like Image Captioning \cite{anderson_bottom-up_2017} or Human-Robot Interaction \cite{fang_understanding_2018}.

One of the greatest difficulties when dealing with visual relations is that the number of possible triplets increases multiplicatively in the cardinality of the human, action and object spaces. Even if we do not distinguish between various ``person'' categories such as ``man'', ``child'' etc., the number of possible interactions -- that is, $ \langle action, object \rangle $ pairs -- grows quadratically. Due to the practical challenges of building a dataset, it is common for only a subset of all possible interactions to be annotated, while a large number remains unlabelled; for instance, HICO \cite{chao_hico:_2015} contains only 600 interactions out of the 9360 possible pairs (among the 9360-600=8760 unlabelled interactions, some are invalid like $ \langle eating, bottle \rangle $, while some are valid but missing like $ \langle carrying, knife \rangle $). This is why more and more approaches are focusing on Zero-Shot Learning (ZSL) for HOI Recognition \cite{shen_scaling_2018,kato_compositional_2018,peyre_detecting_2019,bansal_detecting_2020}. ZSL aims to alleviate the problems caused by the combinatorial growth of the number of possible interactions by allowing models to make predictions about previously unseen interactions.

\begin{figure}[t]
	\centering
	\includegraphics[height=.10\textwidth]{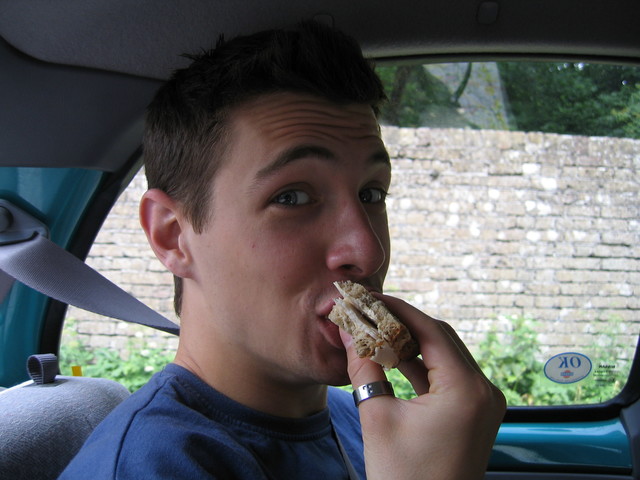}
	~~~
	\includegraphics[height=.10\textwidth]{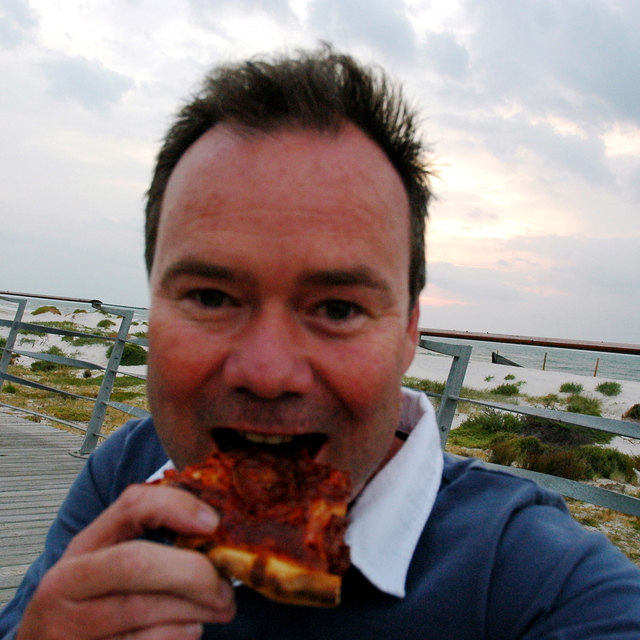}
	~~~
	\includegraphics[height=.10\textwidth]{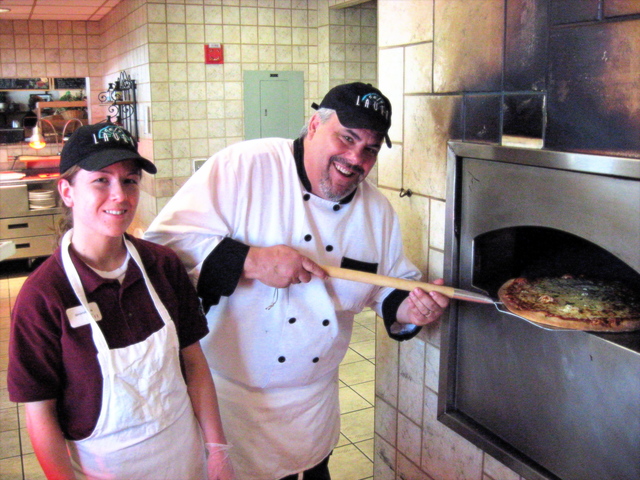}
	\caption{Left to right: $\langle$\textit{eating}, \textit{sandwich}$\rangle$, $\langle$\textit{eating}, \textit{pizza}$\rangle$, $\langle$\textit{cooking}, \textit{pizza}$\rangle$. Both pairs of objects (\textit{pizza} and \textit{sandwich}) and actions (\textit{eating} and \textit{cooking}) are semantically similar, yet images that share an action look more similar than images that share an object.}
	\label{fig:act_sim}
\end{figure}

We focus on actions, as they play a more significant role than objects in defining an interaction: several studies in Psychology \cite{norman_design_2013}, Neurobiology \cite{chao_representation_2000} and Computer Vision \cite{stark_achieving_1991, bansal_detecting_2020} show that objects can be categorised and recognised based on their affordances, making the semantics of an object defined in term of actions, and we empirically verify this intuition via some visual examples provided in \cref{fig:act_sim}. For this reason, we follow a challenging zero-shot setting \cite{kato_compositional_2018} that consists of predicting interactions containing unseen action and object classes, instead of only new combinations of seen classes. We adopt a compositional strategy, as in \cite{shen_scaling_2018,peyre_detecting_2019}: we detect objects and actions first and then combine the results to detect interactions. This is effective in the considered zero-shot setting, where many of the unseen interactions are combinations of a seen object/action with a new action/object, as the model will find it easier to predict the component containing the seen class.

Our model uses a Graph Convolutional Network (GCN) \cite{kipf_semi-supervised_2016} to learn unseen classes in a semi-supervised manner \cite{wang_zero-shot_2018,gao_i_2019}. The graph's connectivity determines how nodes are linked to each other and thus how information is aggregated in the learnt representations. We make use of a particular type of graph called an \textit{affordance graph}, that is, a graph whose edges model \textit{affordances} \cite{norman_design_2013,kato_compositional_2018}: action-object pairs $ \langle a, o \rangle $ where $ a $ can be performed on $ o $ (e.g., $ \langle$\textit{hold}, \textit{apple}$\rangle $, because apples can be held). Such a graph enables the model to learn what interactions are affordable regardless of whether they appear in the training set, allowing it to perform zero-shot predictions.

The focus of this paper is to propose a new training objective function that aims to improve the representations learnt by the model. More specifically, the proposed objective function enhances the loss used by state-of-the-art approaches in two ways. First, it effectively distils action affordance in the unseen class representations by making use of relations from the affordance graph to train unseen actions in a weakly-supervised way. As a result, the model learns to distinguish which unseen actions can be performed on a given object and which ones cannot. Second, it imposes a local structure on the latent space through a regulariser that clusters unseen class representations together with similar classes according to the affordance graph. 
Additionally, we attempt to tackle a shortcoming that affects current approaches: GCN's seen action representations are affected by unseen ones, which are not trained in a fully supervised way and thus add noise. Therefore, we learn an alternative set of representations for seen classes unaffected by unseen ones. Qualitative results demonstrate that our model (shown in \cref{fig:model}) learns representations that are effective at differentiating actions based on affordances, and our experiments show that our model outperforms the current state of the art on HICO \cite{chao_hico:_2015}, VG-HOI \cite{kato_compositional_2018} and COCO-a \cite{ronchi_describing_2015}.

\section{Related Work}
\subsection{Knowledge Usage in HOI Recognition}
Many works have been proposed to perform HOI Detection in recent years, the most similar to ours being the ones that make use of pre-existing knowledge \cite{kato_compositional_2018,peyre_detecting_2019,bansal_detecting_2020,xu_learning_2019}. 
In \cite{bansal_detecting_2020} a language component is used to identify functionally similar objects, effectively augmenting the training data with new interaction instances. In the other works, the pre-existing knowledge is used to obtain class representations, which are used for prediction. %as shown in \cref{fig:zs_framework}. 
These representations come from word embeddings that are mapped through functions implemented as a Multi-Layer Perceptron (MLP) \cite{peyre_detecting_2019} or a GCN \cite{kato_compositional_2018,xu_learning_2019}. An important difference between these models lies in what representations are computed: while in \cite{peyre_detecting_2019} action, object and interactions classes are all considered and the respective scores combined in a compositional way, in the other methods only representations for actions \cite{xu_learning_2019} or interactions \cite{kato_compositional_2018} are used for prediction. Our approach is similar to \cite{peyre_detecting_2019} regarding the compositional model and to \cite{kato_compositional_2018,xu_learning_2019} in the utilisation of external knowledge to build the graph used by the GCN, but differs from all of the above mainly in the way we use the graph at training time to regularise action representation and to distil affordance information into the model.

\subsection{Zero-Shot Learning}
The growing field of ZSL primarily aims to overcome the difficulties of dealing with a non-exhaustively annotated dataset. A common framework to perform ZSL \cite{frome_devise:_2013,zhang_learning_2017,wang_zero-shot_2018,gao_i_2019} is to exploit some kind of pre-existing knowledge to transfer to unseen classes what has been learnt about seen ones in a semi-supervised way. Representations are learnt for both classes and instances and compared through a similarity function to predict output probabilities. The model is trained by feeding the output scores for seen classes into a loss function such as a ranking loss \cite{frome_devise:_2013}, least squares \cite{zhang_learning_2017} or cross entropy \cite{wang_zero-shot_2018,gao_i_2019}.

An interesting method to learn better representations is to add a regularisation loss \cite{mishra_generative_2018,schonfeld_generalized_2018}. In particular, \cite{mishra_generative_2018} maps label embeddings into the visual space, adding a reconstruction loss to make sure that the inverse transformation is also possible and thus the visual projection preserves semantics. A different technique is used in \cite{schonfeld_generalized_2018}, where a cross-reconstruction loss between images and labels is added in order to ``pull together'' representations of the same class from the two different sources (image and labels). Inspired by these works, we formulate a different regularisation loss that uses the affordance graph and is thus better suited to our goal of modelling action affordance.

A few recent approaches tackle ZSL in HOI Recognition/Detection \cite{shen_scaling_2018,kato_compositional_2018,peyre_detecting_2019,bansal_detecting_2020}. We compare our results to the works that considers unseen actions \cite{kato_compositional_2018,peyre_detecting_2019}, as they are the most closely related to ours.

\begin{figure}[t]
	\centering
	
	\subfloat[Train\label{fig:model_train}]{\includegraphics[width=.47\textwidth]{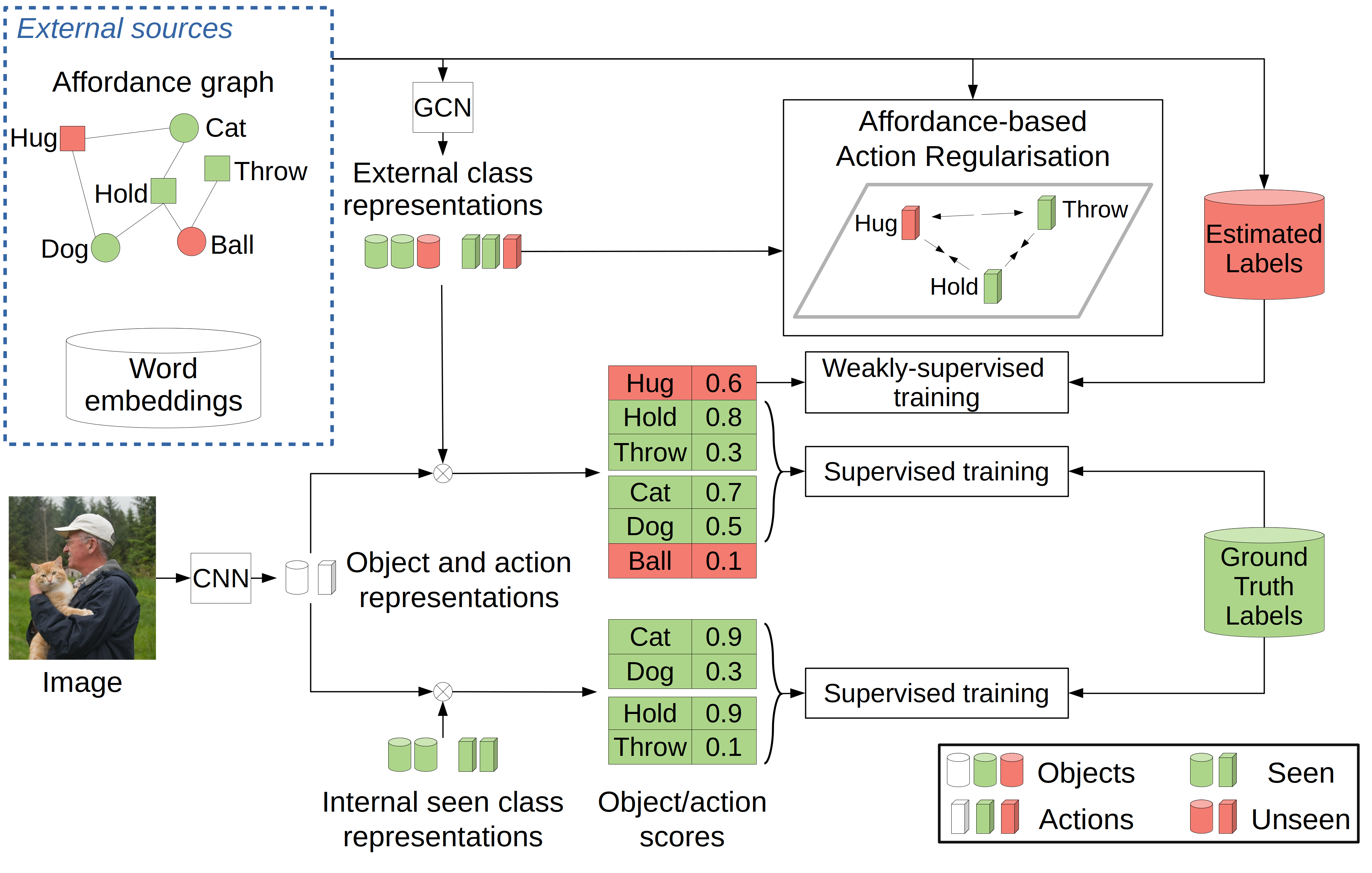}}
	\vspace*{.4cm}
	\subfloat[Inference\label{fig:model_inference}]{\includegraphics[width=.45\textwidth]{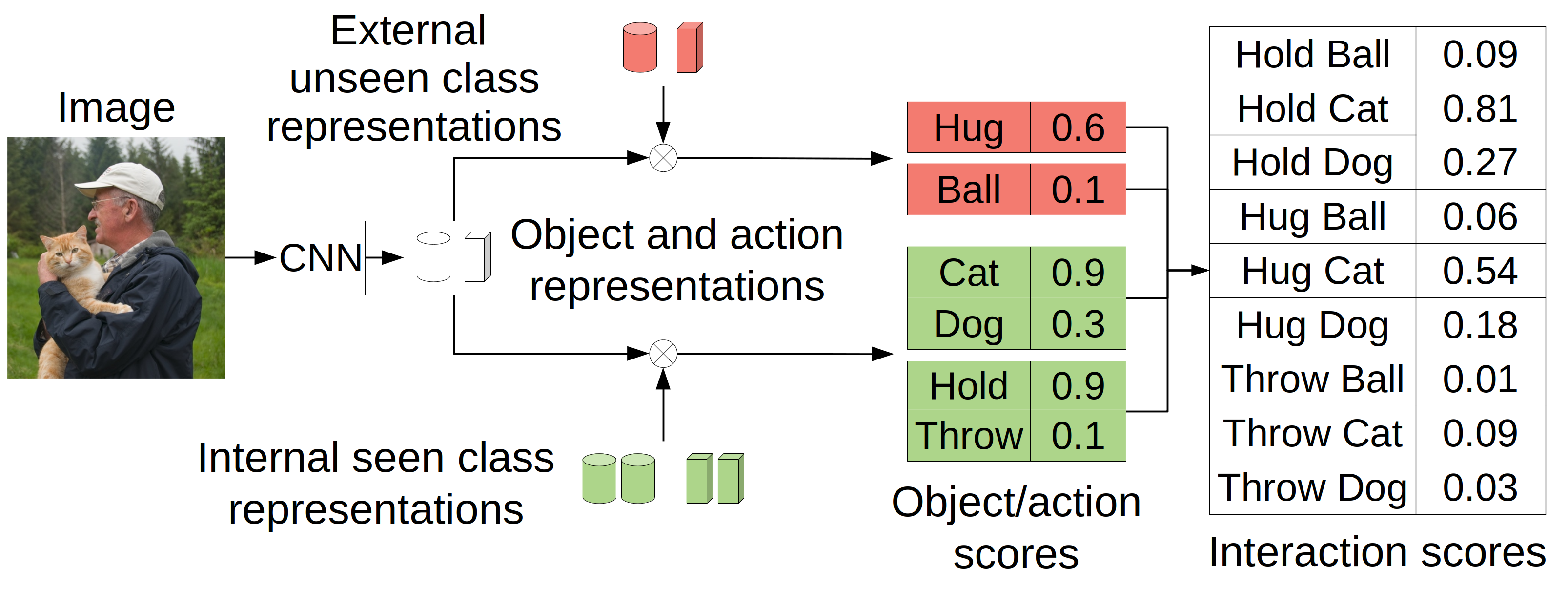}}
	\caption{Overview of the proposed model for HOI Recognition during training \protect\subref{fig:model_train} and inference \protect\subref{fig:model_inference}. $\otimes$ indicates dot product. Best viewed in colour.}
	\label{fig:model}
\end{figure}

\section{Notation and Problem Statement}
Let us denote by $ \mathcal{O} $ and $ \mathcal{A} $ the ordered set of objects and actions, respectively. For instance, we might have \textit{apple} $\in \mathcal{O} $ and \textit{eat} $\in \mathcal{A} $. We will denote the elements of these sets by the corresponding lowercase letter (for example $ o_j $ is the $j$-th element in set $ \mathcal{O} $), or sometimes by the index only (for example we will write $ k \in \mathcal{A}$ instead of $ a_k \in \mathcal{A} $).

Our dataset is denoted by $ \mathcal{D} = \left\{ (I_i, \mathbf{T}_i)\right\}_{i=1}^{M} $. Here, $ I_i $ is the $i$-th image and $ \mathbf{T}_i \in \{0,1\}^{|\mathcal{O| \times |\mathcal{A}|}} $ is its label matrix, with its $jk$-th element $ t_{ijk} $ being 1 if and only if example $ i $ is annotated with interaction $ \langle a_k, o_j \rangle $ (note that an image can have multiple labels). Under the considered Zero-Shot Learning setting, we assume that there are no available visual examples for some objects and actions. This is equivalent to omit the corresponding labels from all images during training, although the affected images might still be annotated with other labels that have not been omitted. The omitted class set will be denoted with $ \mathcal{U} $ (they are \textit{unseen}), while $ \mathcal{S} $ is the set of \textit{seen} (i.e., trained-on) classes. Therefore, we have $ \mathcal{O} = \mathcal{S}^O \cup \mathcal{U}^O $ and $ \mathcal{A} = \mathcal{S}^A \cup \mathcal{U}^A $. Note that seen and unseen classes do not intersect, i.e., $ \mathcal{S}^q \cap \mathcal{U}^q = \emptyset ~ \forall q \in \{O, A\} $. The task is to learn a model that is able to predict any interaction $ \langle a_k, o_j \rangle $, even when $ o_j \in \mathcal{U}^O $ or $ a_k \in \mathcal{U}^A $ (that is, when either or both object and action are unseen).

\section{Proposed Method} \label{sec:method}
\subsection{Affordance Graph}
The main motivation of this work is to improve zero-shot interaction recognition by using structured external knowledge, which is expressed in the form of an \textit{affordance graph}. We define it as a graph $ \mathcal{G} = \langle \mathcal{V}, \mathcal{E} \rangle $ whose nodes $ \mathcal{V} $ are objects and actions and edges $ \mathcal{E} $ represent \textit{affordances} \cite{norman_design_2013}: object node $ o_j $ is connected to action node $ a_k $ only if $ a_k $ can be performed on $ o_j $, i.e., $ \langle a_k, o_j \rangle $ constitutes a valid interaction. For example, \textit{eat} and \textit{apple} will be connected, but \textit{eat} and \textit{fork} will not because people cannot eat forks. This graph is undirected and bipartite: all links are symmetric and there are no connections between object nodes, nor between action nodes. We construct the affordance graph by mining interactions from external sources, to simulate a real-world scenario where no interaction information regarding unseen classes is available. Details about the construction process will be provided in \cref{sec:graph_construction}.

\subsection{Model Architecture} \label{sec:arch}

\subsubsection{Preliminary: Graph Convolutional Networks}
Let us consider a graph with $ N $ nodes, adjacency matrix $ \mathbf{A} $ and initial node representations $ \mathbf{Z}_0 \in \mathbb{R}^{N \times d_0} $ for some dimension $ d_0 $. A single layer of a Graph Convolutional Network (GCN) \cite{kipf_semi-supervised_2016} computes a new representation for each node by aggregating the ones of its neighbours according to $
\mathbf{Z}_1 = \phi_1\big(\tilde{\mathbf{A}}\mathbf{Z}_0\mathbf{\Theta}_1\big) $, where $ \phi $ is an activation function such as ReLU \cite{he_delving_2015}, $ \mathbf{\Theta}_1 \in \mathbb{R}^{d_0 \times d_1} $ are the layer parameters and $\tilde{\mathbf{A}} = \mathbf{D}^{-\frac{1}{2}}(\mathbf{A} + \mathbf{I})\mathbf{D}^{-\frac{1}{2}}$ is the normalised adjacency matrix, where $\mathbf{D} $ is a diagonal matrix with $ d_{ii} = 1 + \sum_{j=1}^{N} a_{ij} $ and $ \mathbf{I} $ is the identity matrix. Deep GCNs can be composed by stacking $ L $ of such layers, producing final representations $ \mathbf{Z} = f_{GCN}(\mathbf{Z}_0) = 
\phi_L\big(\tilde{\mathbf{A}} 
\phi_{L-1}\big( \dots
\phi_1\big(\tilde{\mathbf{A}}\mathbf{Z}_0\mathbf{\Theta}_1 \big)
\dots \big)
\mathbf{\Theta}_L \big) $. 
We refer the reader to \cite{kipf_semi-supervised_2016} for more details.

\subsubsection{Overview} Our model takes as input an image $ I $, which is fed into a Convolutional Neural Network (CNN) such as ResNet \cite{he_deep_2016}, producing image-level visual features $ \boldsymbol{v} = f_{CNN}\left(I\right) $. These features are fed into two identically structured modules indexed by variable $ q $, one for objects ($ q=O $) and one for actions ($ q=A $). Specifically, for each module we compute a $d$-dimensional representation $ \boldsymbol{x}^q = f^q_{1}(\boldsymbol{v}) $ through a non-linear mapping $ f^q_{1} $ (e.g., an MLP). Vector $ \boldsymbol{x}^q $ is compared to a set of $d$-dimensional class representations $ \mathbf{Z}^q = [\boldsymbol{z}^q_1 ~|~ \dots ~|~ \boldsymbol{z}^q_{|\mathcal{S}^q \cup ~\mathcal{U}^q|}] $ through a similarity function $ g(\boldsymbol{x}^q, \boldsymbol{z}_i^q) $, that we implement as inner product following \cite{li_semi-supervised_2015}: $g(\boldsymbol{x}, \boldsymbol{z}) = \boldsymbol{x}^T\boldsymbol{z} $. Similarity scores are fed into the sigmoid function $ \sigma(x) = 1 / (1 + e^{-x})$ to output probabilities $ \boldsymbol{y}^q = \sigma(\mathbf{Z}^q \boldsymbol{x}) $. We will now describe how to compute class representations $ \mathbf{Z}^q $.

\subsubsection{Class Representations}
In our model, akin to \cite{kato_compositional_2018,wang_zero-shot_2018,gao_i_2019}, we use a GCN to train unseen class representations in a semi-supervised way, effectively embedding the affordance relations contained in the graph into the learnt representations. We also incorporate additional semantic information computed from word embeddings, but, differently from previous work, we do not use them to initialise GCN's input embeddings. The reason is that the affordance graph and word embeddings provide different types of semantics: the former aims to capture affordances, while the latter co-occurrence statistics. As a result, for instance, ``eat'' and ``drink'' are  distant according to affordances while close according to word embeddings, which can result in a mismatch in action similarity that brings down the performance (see supplementary). However, co-occurrence semantics carried by word embeddings can be useful for objects (e.g., ``pizza'' and ``sandwich'' have high similarity according to word embeddings, and indeed are both objects that can be eaten), so we use word embeddings to enrich the objects representations produced by the GCN. The final class representations $ \mathbf{Z}_{EXT}^O \in \mathbb{R}^{|\mathcal{O}| \times d} $ and $ \mathbf{Z}_{EXT}^A \in \mathbb{R}^{|\mathcal{A}| \times d} $, that we call \textit{external representations}, are
\begin{align}
&\mathbf{Z}_{EXT}^O = \left(\mathbf{Z}_{GCN}\right)_{\mathcal{O}, :} + f_{2}(\mathbf{W}^O) \\
&\mathbf{Z}_{EXT}^A = \left( \mathbf{Z}_{GCN} \right)_{\mathcal{A}, :} \label{eq:zs_ext_act} \\
&\mathbf{Z}_{GCN}  = f_{3}\left( f_{GCN} \left(\mathbf{Z}_0 \right) \right) ~,
\end{align}
where $ \left(\mathbf{Z}_{GCN}\right)_{\mathcal{O}, :} $ and $ \left(\mathbf{Z}_{GCN}\right)_{\mathcal{A}, :} $ denote the rows of $\mathbf{Z}_{GCN} $ corresponding to object and action classes (respectively), $ f_2 $ and $f_3$ are non-linear functions (e.g., MLPs), $ \mathbf{W}^O \in \mathbb{R}^{|\mathcal{O}| \times d'} $ are $d'$-dimensional word embeddings and GCN's input embeddings $ \mathbf{Z}_0 \in \mathbb{R}^{(|\mathcal{O}| + |\mathcal{A}|) \times d_0} $ are randomly initialised. We use $ \mathbf{Z}_{EXT}^O $ and $ \mathbf{Z}_{EXT}^A $ to predict class probabilities $ \boldsymbol{y}_{EXT}^q = \sigma(\mathbf{Z}_{EXT}^q \boldsymbol{x}) $ for $ q \in \{O, A\} $. Note that these representations (and the corresponding probabilities) are computed for both seen and unseen classes. We use $ \mathbf{Z}_{EXT-\mathcal{S}}^q = \left( \mathbf{Z}_{EXT}^q \right)_{\mathcal{S}^q, :} $ and $\mathbf{Z}_{EXT-\mathcal{U}}^q = \left( \mathbf{Z}_{EXT}^q \right)_{\mathcal{U}^q, :} $ to denote the sub-matrices of $ \mathbf{Z}_{EXT}^q $ that only contain rows for seen or unseen classes, respectively.

The representations computed as above contain informations aggregated from neighbours in the affordance graph. This is why they are well-suited for ZSL, but the downside is that seen class representations are affected by unseen ones. This introduces noise in the representation of seen classes, and in fact we empirically verify that it lowers performance (see supplementary). To overcome this issue, we train an alternative set of representations for seen classes $ \mathbf{Z}_{INT}^q \in \mathbb{R}^{|\mathcal{S}^q| \times d}$, called \textit{internal representation}, in the standard supervised way. This results in separate probability vectors $ \boldsymbol{y}_{INT}^q = \sigma(\mathbf{Z}_{INT}^q \boldsymbol{x}) $ for seen object and action classes.

\subsubsection{Inference} At inference time a score has to be assigned to every interaction, producing a matrix $ \mathbf{Y} \in \left[0, 1\right]^{|\mathcal{O}| \times |\mathcal{A}|} $ whose element $ y_{jk} $ constitutes the probability for interaction $ \langle a_k, o_j \rangle $. We do so by multiplying object and action scores together:
\begin{equation}\arraycolsep=1.0pt\def\arraystretch{1.5}
\begin{aligned}
\mathbf{Y} &= 
\left[\begin{array}{c} \boldsymbol{y}_{INT}^O \\ 
\boldsymbol{y}_{EXT-\mathcal{U}}^O
\end{array}\right] 
\left[\begin{array}{c} \boldsymbol{y}_{INT}^A \\
\boldsymbol{y}_{EXT-\mathcal{U}}^A
\end{array}\right]^T \\
& =
\left[\begin{array}{c} \sigma\left(\mathbf{Z}_{INT}^O \boldsymbol{x} \right) \\ 
\sigma\left(\mathbf{Z}_{EXT-\mathcal{U}}^O \boldsymbol{x} \right)
\end{array}\right] 
\left[\begin{array}{c} \sigma\left(\mathbf{Z}_{INT}^A \boldsymbol{x} \right) \\ 
\sigma\left(\mathbf{Z}_{EXT-\mathcal{U}}^A \boldsymbol{x} \right)
\end{array}\right] ^T ~.
\end{aligned}
\end{equation} 
In our model, external representations of seen classes are used during training to allow for unseen ones to be learnt in a semi-supervised fashion through the GCN, but they are not used for inference (see \cref{fig:model_inference}).

\subsection{Training}
Our model is trained by minimising the following composite loss function, which is designed to optimise both internal class representations $ \mathbf{Z}_{INT}^q $ and all the remaining parameters $ \mathbf{\Theta} $ (which include weights for MLPs and GCN, as well as GCN's initial representations $ \mathbf{Z}_0 $) through variables $ \boldsymbol{y}^q_{INT}$, $\boldsymbol{y}^q_{EXT} $ and $ \mathbf{Z}_{EXT}^q $:
\begin{equation} \label{eq:tot_loss}
\min_{\substack{\mathbf{\Theta} \\ \mathbf{Z}_{INT}^O \\ \mathbf{Z}_{INT}^A \\ }} ~~
\begin{aligned}[t]
&\sum_{i=1}^M \sum_{q \in \{O,A\}} \ell\left( \left(\boldsymbol{y}_{INT}^q\right)_i, \boldsymbol{t}^q_i, \mathcal{S}^q \right) + \\
&\sum_{i=1}^M \sum_{q \in \{O,A\}} \ell\left( \left(\boldsymbol{y}_{EXT-\mathcal{S}}^q\right)_i, \boldsymbol{t}^q_i, \mathcal{S}^q \right) + \\
&\sum_{i=1}^M \lambda ~ \ell\left(\left(\boldsymbol{y}^A_{EXT-\mathcal{U}}\right)_i, \hat{\boldsymbol{t}}^A_i \hspace*{.3mm} , \mathcal{U}^A \right) + \\
&~\rho\mathcal{L}_{REG}\left(\mathbf{Z}_{EXT}^A \right) ~, \\
\end{aligned}
\end{equation}
where $ \lambda $ and $ \rho $ are hyperparameters that regulate the contribution of their respective terms, label vectors $ \boldsymbol{t}_i^O \in \{0,1\}^{|\mathcal{O}|} $ and $ \boldsymbol{t}_i^A \in \{0,1\}^{|\mathcal{A}|} $ are obtained from matrix $ \mathbf{T}_i $ according to $ t_{ij}^O = \max_{k \in \mathcal{A} } t_{ijk} $ and $ t_{ik}^A = \max_{j \in \mathcal{O} } t_{ijk} $, and $ \ell $ is the standard binary cross entropy loss:
\begin{equation} \label{eq:bce}
\ell\left(\boldsymbol{y}, \boldsymbol{t}, \mathcal{J} \right) = \sum_{j \in \mathcal{J}} \left[t_{j} \log y_{j} + (1 - t_{j}) \log (1 - y_{j}) \right] ~,
\end{equation}
where $ \boldsymbol{y} $ are outputs, $ \boldsymbol{t} $ target labels and $ \mathcal{J} $ a set of indices. We also add $L_2$-regularisation to $ \mathbf{\Theta} $ to prevent overfitting (not shown in \cref{eq:tot_loss}).

The first two terms of \cref{eq:tot_loss} implement a standard training loss, which uses ground truth labels to reward pairing instances with the corresponding seen classes and to penalise assigning the wrong class. The third term aims to train unseen actions in the same way. However, since ground truth labels are not available for unseen actions, we adopt a weakly-supervised approach and estimate labels $ \hat{\boldsymbol{t}}^A $ as:
\begin{gather}
\hat{t}^A_{k} = \max_{j \in \mathcal{S}^O} m_{jk}s_{jk}  \qquad \forall k \in \mathcal{U}^A \label{eq:loss_dis_t_hat} \\
s_{jk} = \dfrac{1}{\sum_{h \in \mathcal{S}^A} t_{jh}} \sum_{h \in \mathcal{S}^A} t_{jh} \left[\boldsymbol{w}_h^T \boldsymbol{w}_k\right]_+ \label{eq:loss_dis_s} ~,
\end{gather}
where $ [x]_+ = \max(x, 0) $, $ \mathbf{M} \in \{0, 1\}^{|\mathcal{O}| \times |\mathcal{A}|} $ is the graph adjacency matrix\footnote{$ \mathbf{M} $ does not need to be a square matrix because the graph is bipartite.} and $ \boldsymbol{w}_k $ is the word embedding for the $k$-th action. 
\cref{eq:loss_dis_s} computes a score that determines how likely unseen action $ k $ describes an image containing object $ j $. This score is \textit{not} binary, but rather a real value in $ [0, 1] $. This is needed because binary estimated labels would incur the risk of introducing noise, since we cannot know which of the affordable unseen actions are actually depicted in a particular image. Word embeddings are used to assign a score based on the similarity with labelled seen actions (which are compatible with object $ j $, since they come from the ground truth) through the positive inner product $ \left[\boldsymbol{w}_h^T \boldsymbol{w}_k\right]_+ $, so that unseen actions similar to shown seen ones will be assigned a higher score: if $ o_j =$ \textit{person} and \textit{hug} is a labelled seen action, \textit{kiss} and \textit{greet} are better unseen candidates than \textit{teach}. Action affordance is distilled into the model according to \cref{eq:loss_dis_t_hat}: score $ s_{jk} $ contributes to $ \hat{t}^A_{k} $ only if $ m_{jk} = 1 $, that is, only if $ \langle a_k, o_j \rangle $ is an affordable action. Since an image may contain multiple objects, the maximum score over objects is taken according to the Multiple Instance Learning framework \cite{mallya_learning_2016}.

Additionally, we use the affordance graph as a regulariser for action classes, with the goal of learning better representations by inducing a structure onto the latent space based on affordances. Specifically, we want to group \textit{functionally} similar actions, that is, actions that can be performed on the same objects. To this end, we use the following ranking margin loss:
\begin{equation} \label{eq:loss_reg} 
\begin{gathered}
\mathcal{L}_{REG}\left(\mathbf{Z}_{EXT}^A \right) = \sum_{i \in \mathcal{U}^A} \sum_{j \in \mathcal{N}(i)} \sum_{k \not\in \mathcal{N}(i)} \left[ \gamma - c_{ij} + c_{ik} \right]_+ \\
c_{ij} = \dfrac{\boldsymbol{z}_i^T\boldsymbol{z}_j}{||\boldsymbol{z}_i|| ||\boldsymbol{z}_j||} \quad \forall i,j \in \mathcal{A} ~,
\end{gathered}
\end{equation}
where $ \gamma \in \mathbb{R} $ is the margin, $ c_{ij} $ is the cosine similarity between the $ i $-th and $ j $-th columns of $ \mathbf{Z}_{EXT}^A $ ($ \boldsymbol{z}_i $ and $ \boldsymbol{z}_j $), and $ \mathcal{N}(i) $ denotes the set of actions that are functionally similar to action node $ a_i $ (i.e., actions at distance 2 from $a_i$ in the affordance graph).

We train our model using Stochastic Gradient Descent (SGD) with momentum \cite{rumelhart_learning_1988} and a fixed learning rate. Further details will be provided in \cref{sec:impl_details}.

\section{Experiments}
We compare our results to the methods reported in \cite{kato_compositional_2018} on HICO and VG-HOI.
Although our work is focused on HOI Recognition (\cref{sec:exp_settings} and \ref{sec:recognition_results}), we also consider the Detection task (\cref{sec:detection_res}), in which the model is required to localise each prediction. We perform Detection experiments on both HICO-DET and COCO-a.

\subsection{Datasets}

\subsubsection{HICO and HICO-DET} The HICO dataset \cite{chao_hico:_2015} and its bounding-box-annotated variant HICO-DET \cite{chao_learning_2018} comprise ~47k images, COCO's 80 object classes \cite{lin_microsoft_2014}, and 117 action classes, including a null one. They are annotated with 600 interactions and each image may belong to more than one interaction class. We follow the predefined train/test split of 38,116/9,658 images. Furthermore, we randomly sample 10\% of the training set for validation in every run. In our Recognition experiment we follow \cite{kato_compositional_2018}, excluding the null action during training and testing and thus restricting the dataset to 116 actions and 520 interactions.

\subsubsection{VG-HOI}
VG-HOI \cite{kato_compositional_2018} is a dataset for Human-Object Interaction built out of Visual Genome \cite{krishna_visual_2017}. It comprises 10,799 train images and 4251 test images, for a total of 15,050. We use 10\% of the training set for validation. There are 1392 objects, 495 actions and 6643 interactions, although for testing only the 532 that have at least 10 instances are used. The much larger number of classes (compared to HICO), together with the lower number of examples, make this dataset extremely challenging.

\subsubsection{COCO-a} COCO-a \cite{ronchi_describing_2015} contains 4413 images annotated with 145 action classes and 80 object classes (same as COCO and HICO), for a total of 1681 interactions. We use it as an evaluation dataset for our model trained on HICO-DET, following the challenging setting used in \cite{peyre_detecting_2019}.

\subsection{Affordance Graph Construction} \label{sec:graph_construction}
To build the affordance graph, we mine interactions from external knowledge bases and add them to the ones that can be found in the training set. Specifically, we use four external sources: Visual Genome \cite{krishna_visual_2017} (except for VG-HOI), ActivityNet Captions \cite{krishna_dense-captioning_2017}, imSitu \cite{yatskar_situation_2016} and HCVRD \cite{zhuang_hcvrd:_2018}. The former three contain image or video captions that we parse into action-object pairs using NLTK \cite{bird_natural_2009} and the dependency parser from AllenNLP \cite{gardner_allennlp:_2018}. On the other hand, HCVRD is annotated with triplets in the form $ \langle subject, predicate, object \rangle $. We select the ones where the $ subject $ is a person and $ predicate $ is an action. Note that, in all cases, we do not add extra nodes into our graph and instead discard interactions containing actions or objects not in $ \mathcal{A} $ and $ \mathcal{O} $ (respectively).

\subsection{Experimental Setting} \label{sec:exp_settings}

\subsubsection{Compared Models} We use four variants of our model: our baseline ($ \rho = 0$ and $ \lambda = 0 $) and the models obtained by only adding one of the proposed loss components ($ \rho > 0$ or $ \lambda > 0 $) or adding both ($ \rho > 0$ and $ \lambda > 0 $). 

The most similar method to ours is \cite{kato_compositional_2018}, which performs zero-shot learning on both action and objects. We compare our models to their best results, which are denoted by ``GCNCL'' followed by different endings based on how the knowledge graph is built. We also report other competitive methods from \cite{kato_compositional_2018}, namely Semantic Embedding Space (SES, \cite{xu_semantic_2015}) and Triplet Siamese. We refer the reader to the corresponding papers for more details.

\subsubsection{Evaluation} We use the standard mean Average Precision (mAP) as evaluation metric, reporting it as a percentage. We train every model multiple times (10 for HICO and 5 for VG-HOI), reporting the average result on the test set. We run Student's t-tests against current state-of-the-art results and all reported improvements are statistically significant at the 99\% confidence interval.

\subsubsection{Zero-Shot Settings} In order to make a fair comparison, we use the same seen/unseen splits as Task 2 from \cite{kato_compositional_2018}: the training set is made of 49 objects and 53 actions for HICO and 554 objects and 198 actions for VG-HOI. At test time all classes are included, following the \textit{Generalised} Zero-Shot Learning setting.

\subsubsection{Implementation Details} \label{sec:impl_details}
We use a ResNet-152 pre-trained on ImageNet \cite{deng_imagenet:_2009} as image feature extractor (same as \cite{kato_compositional_2018}). Functions $ f_{1} $, $ f_{2} $ and $ f_{3} $ are implemented by two fully-connected layers with output dimensions both equal to 1024, with ReLU non-linearity. After the non-linearity we add Dropout \cite{hinton_improving_2012} (at a 0.5 rate) for $ f_{1} $ and $ f_{3} $, but not for $ f_{2} $, as suggested in \cite{peyre_detecting_2019}. We use Glorot initialisation \cite{glorot_understanding_2010} to initialise the optimisation parameters in \cref{eq:tot_loss}. Our GCN comprises two convolutional layers with output dimension 1024, the first of which is equipped with ReLU and Dropout (0.5 rate).

We keep the margin parameter $ \gamma $ in \cref{eq:loss_reg} fixed at $0.3$, whereas we experiment with different values of $ \rho $ and $ \lambda $ for the two datasets. The best ones (according to validation results) are the ones shown in the respective tables.

We use GloVe \cite{pennington_glove:_2014} for our word embeddings. More specifically, we use the 300-dimensional embeddings trained on Gigaword and Wikipedia\footnote{Available at \url{https://nlp.stanford.edu/projects/glove}.} and we normalise them. For compound words, we take the average of the components.

Finally, we train our model using minibatch Stochastic Gradient Descent (SGD) with momentum. We use a fixed learning rate of  $ 0.001 $ and set the momentum and weight decay coefficients to $0.9$ and $ 5 \cdot 10^{-4} $, respectively. We train our model for a maximum of 100 epochs on HICO and 150 on VG-HOI, with early stopping based on validation accuracy. We use a batch size of 64.

\subsection{Results} \label{sec:recognition_results}

\begin{table}[t]
	\centering
	\begin{tabular}{lrr}
		Method	 & 	          All & 		Unseen only \\
		\hline
		Triplet Siamese&  10.38  & 7.76	\\
		SES & 11.69  & 7.19	\\
		GCNCL-I & 11.93  & 7.22   \\
		GCNCL+NV+A & 11.94  & 7.50	\\
		\hline
		Ours	                                 	 & 		   13.79  &              6.93   \\
		Ours, $ \lambda = 1 $            	 & \textbf{16.02} &     		10.08   \\
		Ours, $ \rho = 10 $          	 &         14.02  &              7.16   \\
		Ours, $\lambda=1,\rho=10$	 & \textbf{16.02} &     \textbf{10.20}  \\
	\end{tabular}
	\caption{Results on HICO.}
	\label{tb:res_hico}
\end{table}

\begin{table}[t]
	\centering
	\begin{tabular}{lrr}
		Method 	  &   		   All &       Unseen only   \\
		\hline
		Triplet Siamese &	    	 2.55  &			  1.67   \\
		SES   &	   		 2.07  &			  0.96   \\
		GCNCL-I+A  &     	 4.00  &              2.63   \\
		GCNCL+A  &     	 4.07  &              2.44   \\
		\hline
		Ours	                                 	  &		     4.90  &              3.51   \\
		Ours, $ \lambda = 0.1 $          	  &   		 5.09  &      		  3.77  \\
		Ours, $ \rho = 100 $          	  &  \textbf{5.11} &      \textbf{3.90}  \\
		Ours, $\lambda=0.1,\rho=100$ &        	 5.01  &              3.74   \\
	\end{tabular}
	\caption{Results on VG-HOI.}
	\label{tb:res_vghoi}
\end{table}

\subsubsection{Results on HICO} 
Our results are summarised in \cref{tb:res_hico}. 
We see that our baseline model already compares very favourably to all the existing approaches, and adding either or both of the proposed losses upgrades our baseline's performance considerably. The best performing model, obtained with $\lambda=1 $ and $\rho=10$, gains more than 4\% over the current state of the art (GCNCL+NV+A) for the whole test set and around 2.7\% for unseen classes only. This corresponds to sizeable $\sim$35\% relative increases.
It is worth mentioning that the graph building process results in 68 missing interactions out of HICO's 520, since they cannot be mined from our external sources. Despite this, no object is completely isolated in the affordance graph, whereas only 5 actions are (\textit{hop\_on}, \textit{hunt}, \textit{lose}, \textit{stab}). Most of these actions (namely \textit{hunt}, \textit{lose}, \textit{stab} and \textit{toast}) are too niche to be found in the other sources, and in fact even in HICO they only appear in one interaction each. On the other hand, \textit{hop\_on} can be found, but not with the meaning of ``jumping on a ride'' it has in HICO (and thus it is not paired with the same objects).
Nonetheless, our model still performs very well, possibly due to the fact that additional interactions are added and they contribute to meaningful representations being learnt, even though they do not appear in HICO.

\begin{figure}[t]
	\centering
	\includegraphics[width=.45\textwidth]{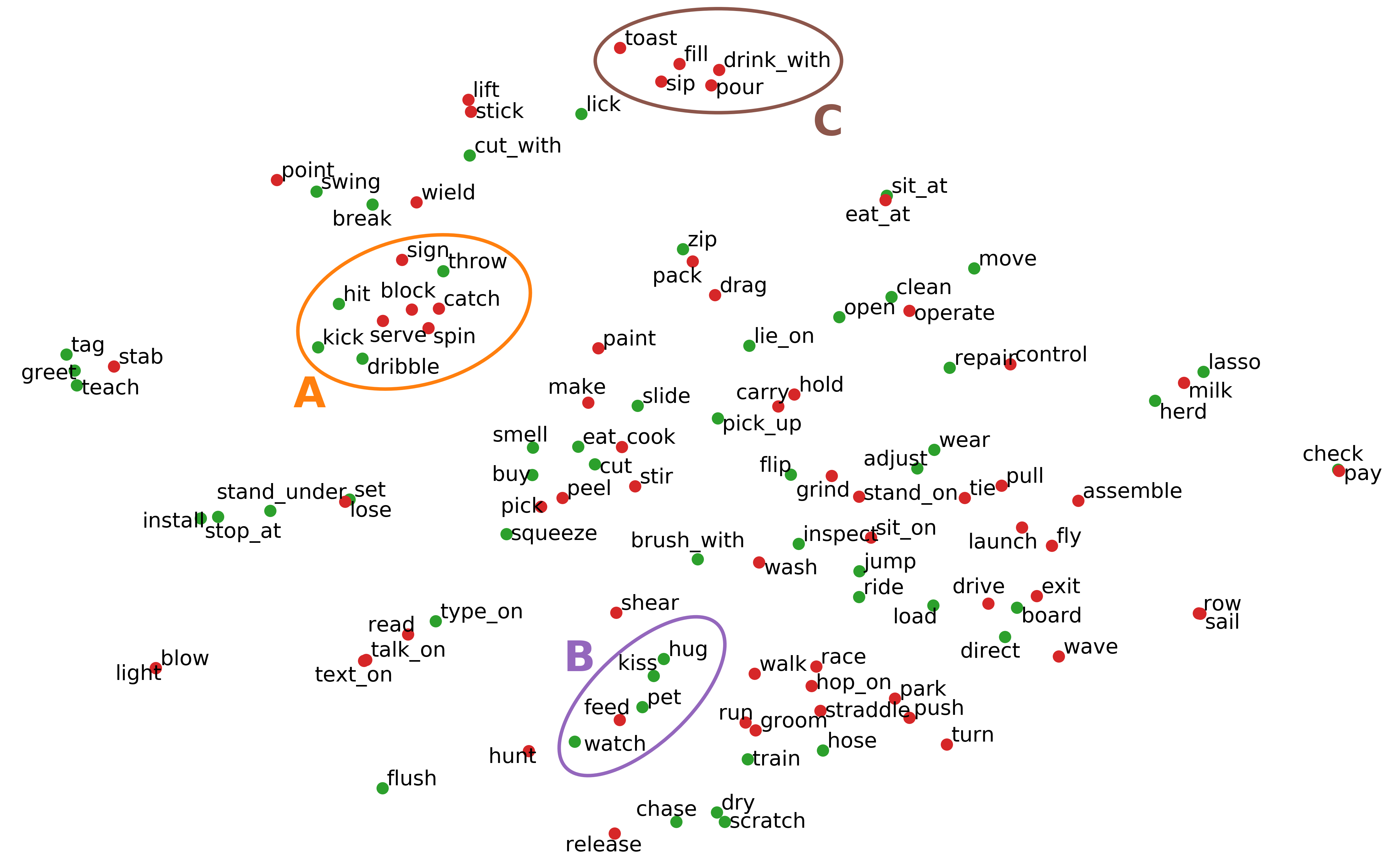}
	\caption{Visualisation of action class representations. Green dots represent seen actions and red dots unseen ones. Some clusters are highlighted: (\textit{A}) sports actions (e.g., \textit{catch},  \textit{throw}), (\textit{B}) actions regarding domesticated animals (e.g., \textit{pet}, \textit{feed}) and (\textit{C}) action involving cups or glasses (e.g., \textit{sip}, \textit{pour}).}
	\label{fig:tsne}
\end{figure}

\subsubsection{Results on VG-HOI} Results are reported in \cref{tb:res_vghoi}. Our baseline is better than previous models, GCNCL+A in particular: $\sim$.83\% for all classes ($\sim$20\% relative gain) and almost 1.1\% for unseen categories, corresponding to a remarkable 40\% relative improvement. Adding the proposed losses improves performance, with the best one obtained by setting $ \lambda=0, \rho = 100 $. While our losses improve results, they are not as effective on this dataset as on HICO. We believe this can be ascribed to the vast number of unseen categories: while in HICO there are 80 objects and 116 actions, VG-HOI contains $\sim$17 times as many objects and more than 4 times as many actions. The sheer number of unseen classes makes classification much more difficult; in particular, our method relies on seen object labels to estimate unseen action ones (see \cref{eq:loss_dis_t_hat}), therefore missing a large amount of information about objects is detrimental. The incompleteness of the affordance graph is also likely to negatively affect performance, as the graph only covers 2753 interactions ($\sim$41\%), 291 actions ($\sim$59\%) and 806 objects ($\sim$58\%). Despite these difficulties, our method still performs significantly better than previous approaches.

\begin{figure}[t]
	\captionsetup{justification=centering}
	\centering
	\subfloat[][\\ \cmark \textit{eat\_at} dining\_table \\ \cmark sit\_at dining\_table]{\includegraphics[height=.15\textwidth]{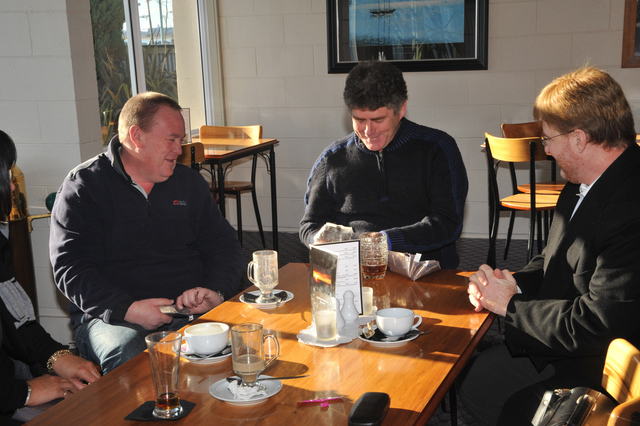}}
	\hspace*{.1cm}
	\subfloat[][\\ \cmark \textit{hold} book \\ \cmark open book]{%
		\includegraphics[height=.15\textwidth]{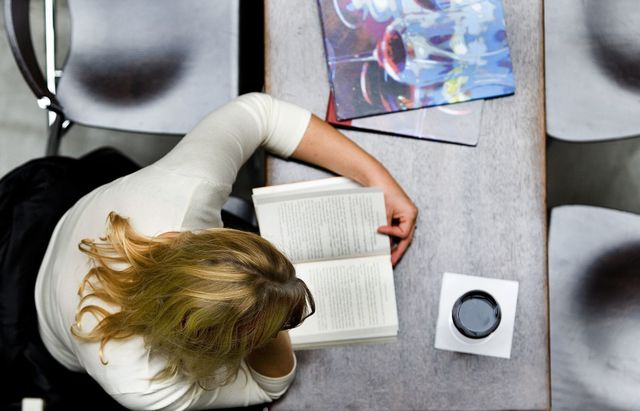}}
	\hspace*{.1cm}
	
	\subfloat[][\\ \cmark \textit{carry} umbrella \\ \cmark \textit{hold} umbrella \\ \xmark stand\_under umbrella]{%
		\includegraphics[height=.15\textwidth]{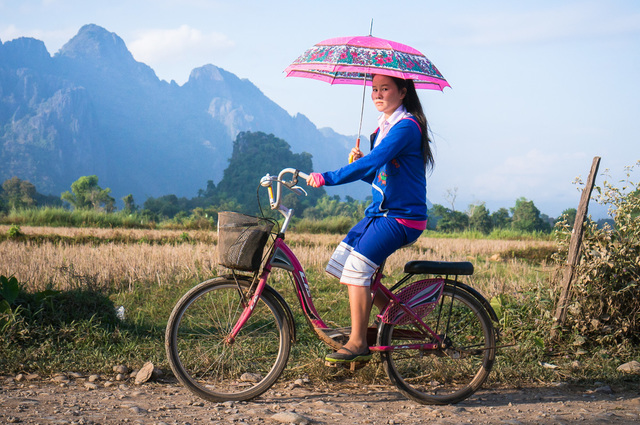}}
	\hspace*{.1cm}
	\subfloat[][\\ \cmark \textit{pull} tie \\ \cmark \textit{tie} tie \\ \cmark wear tie]{%
		\includegraphics[height=.15\textwidth]{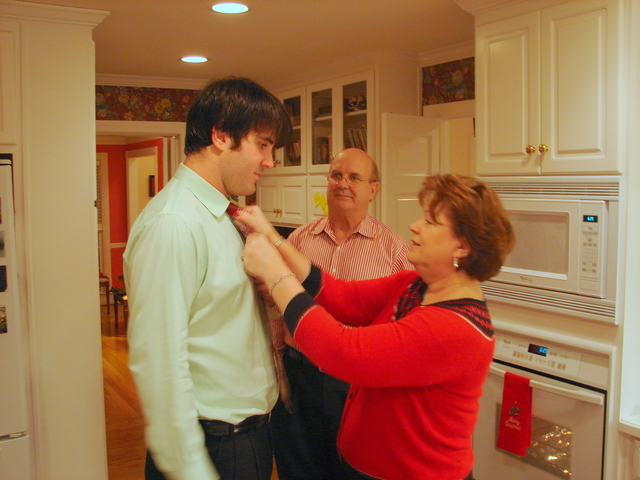}}
	\captionsetup{justification=justified}
	\caption{Some predictions of our best model on HICO. Marks indicate whether the prediction matches the ground truth \cmark or not \xmark. Actions in \textit{italic} are unseen.}
	\label{fig:qual}
\end{figure}

\subsubsection{Qualitative Results on HICO}
We show some predictions on HICO's test set examples in \cref{fig:qual}, demonstrating that our model is able to correctly predict several previously unseen actions. We also show the representation space in \cref{fig:tsne} using t-SNE \cite{maaten_visualizing_2008} on a model trained with both proposed losses ($\lambda, \rho > 0$). Some clusters are clearly identifiable, such as cluster A, which contains actions such as \textit{catch}, \textit{throw} or \textit{spin} that can be performed on small sport items like \textit{sports\_ball} or \textit{frisbee}. This shows that the proposed approach is effective in grouping actions based on their affordance. Comparisons between representation spaces obtained in different settings can be found in the supplementary material.

\begin{table}[t]
	\centering
	\begin{tabular}{lr}
		Method 	& All\\
		\hline
		Shen \textit{et al.} \cite{shen_scaling_2018} & 6.46 \\
		Chao \textit{et al.} \cite{chao_learning_2018}	&	 7.81 \\
		InteractNet \cite{gkioxari_detecting_2018} & 9.94 \\
		GPNN \cite{qi_learning_2018}	&   13.11 \\
		Xu \textit{et al.} \cite{xu_learning_2019}  	&   14.70 \\
		iCAN \cite{gao_ican:_2018}	&   14.84 \\
		Song \textit{et al.} \cite{song_novel_2020} & 15.27 \\
		Wang \textit{et al.} \cite{wang_deep_2019} &   16.24 \\
		No-frills \cite{gupta_no-frills_2019} &   17.18 \\
		Li \textit{et al.} \cite{li_transferable_2019} &   17.22 \\
		RPNN \cite{zhou_relation_2019} 	&   17.35 \\
		PMFNet \cite{wan_pose-aware_2019}	&   17.46 \\
		Peyre \textit{et al.} \cite{peyre_detecting_2019} & 19.40\\
		Wang \textit{et al.} \cite{wang_learning_2020} & 19.56 \\
		PPDM \cite{liao_ppdm_2020} & 21.73 \\
		Bansal \textit{et al.} \cite{bansal_detecting_2020} & \textbf{21.96}  \\
		\hline
		Ours   				&  18.74\\
	\end{tabular}
	\caption{Results on HICO-DET in a fully supervised setting.}
	\label{tb:hicodet_res}
\end{table}

\begin{table}[t]
	\centering
	\begin{tabular}{lrr}
		Method							 				& 			  All &            Unseen  \\
		\hline	
		Ours	& 11.18  & 8.19   \\
		%			Ours, $\lambda=0.1$	& 11.86  & 9.57   \\
		Ours, $\lambda=0.1, \rho=0.1$	& \textbf{11.94}  & \textbf{9.81}   \\
	\end{tabular}
	\caption{Baseline for ZS HOI Detection on HICO-DET.} 
	\label{tb:hicodet_zs_res}
\end{table}

\subsection{Zero-Shot HOI Detection} \label{sec:detection_res}
We used different settings for the Detection experiments, which can be found in the supplementary material. We present the results in the following.

\subsubsection{HICO-DET}
While there are works on Zero-Shot Learning on HICO-DET for interactions \cite{shen_scaling_2018,peyre_detecting_2019} and objects \cite{bansal_detecting_2020}, no previous approach has dealt with zero-shot actions (to the best of our knowledge). We provide in \cref{tb:hicodet_zs_res} a baseline for future reference.
%, in line with recent works \cite{xu_learning_2019,bansal_detecting_2020}. 
We also show in \cref{tb:hicodet_res} how our approach compares against other methods in a fully supervised setting as a reference, where we can see that there is a noticeable increase in mAP with respect to most methods in the literature. It is worth mentioning that some of the techniques that likely contribute to the outstanding results of the top three methods, such as fine-tuning the object detector on HICO-DET \cite{bansal_detecting_2020} or following a more intensive training regime while fine-tuning the feature extractor (50 epochs on 5 GPUs for \cite{wang_learning_2020}, 110 epochs on 8 GPUs for \cite{liao_ppdm_2020}), are applicable to our model as well -- in fact, Bansal \textit{et al.} report that their method only achieves 16.96\% mAP without such fine-tuning. We leave this for future work. 

\subsubsection{COCO-a}
In \cref{tb:cocoa_zs_res} we compare our results on COCO-a (reporting our baseline plus the best hyperparameter setting for each column) against a state-of-the-art approach using the challenging setting described in their paper: train on HICO-DET and evaluate on COCO-a. Under this setting, there are 1474 unseen interactions, 1048 of which involve an unseen action. Our approach performs much better than the best one from \cite{peyre_detecting_2019}, gaining around 2.7 points for all unseen interactions ($\sim$40\% relative gain) and 3.7 points when dealing with interactions involving unseen actions (about 50\% relative gain). Performance improve even further when setting $ \lambda $ and/or $ \rho $ to non-zero values (the best assignment for each measure is reported).

\begin{table}[!t]
	\centering
	\begin{tabular}{lrr}
		& \multicolumn{2}{c}{Unseen HOIs} \\
		Method & All & With unseen action  \\
		\hline
		Peyre \textit{et al.} \cite{peyre_detecting_2019} (best) & 6.9 & 7.3   \\
		Ours & 9.65 & 11.00   \\
		Ours, $\lambda=0.1$ & 9.93 & \textbf{11.44} \\
		Ours, $\lambda=0.1,\rho=10$ & \textbf{10.01} & 11.13 \\
	\end{tabular}
	\caption{Results on COCO-a.} 
	\label{tb:cocoa_zs_res}
\end{table}

\section{Conclusion}
We have proposed an effective approach that uses structured knowledge in the form of an affordance graph to improve Zero-Shot Human-Object Interaction Recognition. The proposed model learns regularised representations of unseen classes in a weakly supervised way using labels which are estimated through the affordance graph, while simultaneously learning representation of seen classes in a supervised fashion. Our method is able to predict unseen interactions in the very challenging case where only about half of the object and action classes are seen during training. We evaluate our results on several datasets (including standard benchmarks like HICO and HICO-DET) and show that our approach performs significantly better than the current state of the art.

\bibliographystyle{ieeetr}
\bibliography{library}

\appendix

\section{Supplementary Material}
This supplementary material contains:
\begin{enumerate}
	\item Details about the hardware and software infrastructure used to implement the method;
	\item Several ablation studies (\cref{sec:ablation}) and additional experiments (\cref{sec:sensitivity});
	\item Details about the HOI Detection experiment described in \cref{sec:detection_res} (\cref{sec:details_hoid});
	\item A visual comparison of HICO's action representation spaces obtained by different means (\cref{sec:vis}) to expand on what we show in \cref{sec:recognition_results}.
\end{enumerate}

\subsection{Infrastructure Details}
The model has been implemented in Python 3.6 using PyTorch v0.4.1. Experiments have been run on a single NVIDIA GeForce GTX TITAN X GPU on a server with an Intel(R) Core(TM) i7-5930K CPU and 64GB of RAM running CentOS Linux 7. The code will be made available upon publication. 

\subsection{Ablation experiments} \label{sec:ablation}
In this section we describe several ablation experiments. All the reported results are computed on HICO.

\subsubsection{Alternative representation}
As described in \cref{sec:arch}, we train an alternative representation for seen classes that we call \textit{internal representation}, denoted by $ \mathbf{Z}^q_{INT} $ for $ q \in \{O, A\} $. The rationale for this is that the semi-supervised training used to train representation of unseen classes via the GCN might introduce noise in the representation of seen classes, which could be trained in a fully-supervised fashion thank to the availability of instance labels. Results shown in \cref{tb:abl_int} corroborate our hypothesis: models that learn the internal representation for seen classes perform consistently better than the corresponding ones which only learn GCN representations. All results are statistically significant at the 95\% confidence interval.

\subsubsection{Word embeddings}
In \cref{sec:arch} we argued that word embeddings are not well-suited to provide affordance information about actions, because word embeddings relate words based on co-occurrence in a sentence. Therefore, while they can capture affordance-based similarity for objects (e.g., in the sentence ``I eat an apple and a banana'', objects \textit{apple} and \textit{banana} co-occur because they both afford action \textit{eating}), this effect is weaker for actions (e.g., in the sentence ``People were eating and drinking'' the actions co-occur not because they are afforded by the same objects, but rather because they can be performed in the same context). To empirically verify this intuition, we perform an ablation experiment whose result are shown in \cref{tb:abl_wemb}. Differences are statistically significant at the 99\% confidence interval. These results justify why we do not add a component based on word embeddings in \cref{eq:zs_ext_act}.

\begin{table}[t]
	\centering
	\begin{tabular}{l|rr|rr}
		\multicolumn{1}{l}{~} & \multicolumn{2}{|c}{With} & \multicolumn{2}{|c}{Without}  \\
		Method						 				 & 			  All &   	       Unseen   
		&  		  All &    		   Unseen   \\
		\hline
		$ \lambda = \rho = 0$       	 & 		   13.79  &              6.93   
		&         13.11  &              6.75   \\
		
		$ \lambda = 1 $            			 &         16.02  &     		10.08   
		&  	   15.14  &     		 9.85   \\
		
		$ \rho = 10 $          				 &         14.02  &              7.16   
		&         13.32  &              6.94   \\
		
		$\lambda=1,\rho=10$	 				 &         16.02  &             10.20   
		&         15.14  &              9.95   \\
	\end{tabular}
	\caption{Ablation: internal representation for seen classes.}
	\label{tb:abl_int}
\end{table}

\begin{table}[t]
	\centering
	\begin{tabular}{l|rr|rr}
		\multicolumn{1}{l}{~} & \multicolumn{2}{|c}{Objects only} & \multicolumn{2}{|c}{Objects and actions} \\
		Method						 				 & 			  All &   	       Unseen   
		&  		  All &    		   Unseen   \\
		\hline
		$ \lambda = \rho = 0$	                                 	 & 		   13.79  &              6.93   
		&         13.31  &              6.20   \\
	\end{tabular}
	\caption{Ablation: word embeddings.}
	\label{tb:abl_wemb}
\end{table}

\subsection{Sensitivity Experiments} \label{sec:sensitivity}
In this section we evaluate how sensitive our model is to the available information, in particular to the amount of unseen labels and completeness of the affordance graph.

\subsubsection{Amount of unseen labels}
We perform an experiment to evaluate how much the amount of unseen labels impacts performance. Specifically, we define a hyperparameter $ \mu \in [0, 1] $ as the ratio of unseen classes with respect to the experiment reported in \cref{tb:res_hico}. Thus, a $ \mu = 1 $ corresponds to the reported experiment (31 unseen object classes and 63 unseen action classes), $ \mu = 0.6 $ means to keep around 60\% of the unseen classes (for a total of 19/38 unseen object/action classes, while the remaining 40\% are added to the seen classes) and a value of$ \mu=0.3 $ means to only keep around 30\% of the unseen classes. Results are shown in \cref{tb:abl_unseen} and show that the amount of unseen labels greatly affects performance.

\begin{table}[t]
	\begin{center}
		\begin{tabular}{lrr}
			Method						 				&            		All 	& Unseen  \\
			\hline
			$\lambda=1$, $ \mu=1 $ (\cref{tb:res_hico})	&  		 16.02	& 10.08 \\
			$\lambda=1$, $ \mu=0.6 $			& 		 20.91 & 12.72 \\
			$\lambda=1$, $ \mu=0.3 $			&\textbf{27.19}	& \textbf{18.21} \\
		\end{tabular}
		\caption{Sensitivity of the model to the amount of unseen labels.}
		\label{tb:abl_unseen}
	\end{center}
\end{table}

\subsubsection{Completeness of the affordance graph}
Our approach makes extensive use of the affordance graph. It is natural to assume that a sparser graph leads to worse results, since it contains less information, thus we design an experiment to verify this assumption. In particular, we define a hyperparameter $ \nu \in [0, 1] $ as the proportion of edges of the affordance graph with respect to one used in the experiments reported in \cref{tb:res_hico}. For instance, $ \nu = 0.8 $ means to sample around 80\% of the edges to keep and remove the remaining 20\%. Results can be viewed in \cref{tb:abl_graph} and they show that the more sparse the affordance graph, the lower the performance.

\begin{table}[t]
	\begin{center}
		\begin{tabular}{lrr}
			Method						 				&       	   		 	All	& Unseen  \\
			\hline
			$\lambda=1$, $ \nu=1 $ (\cref{tb:res_hico})	&\textbf{16.02}	& \textbf{10.08} \\
			$\lambda=1$, $ \nu=0.8 $ & 15.43&  9.34 \\
			$\lambda=1$, $ \nu=0.6 $ & 14.50&  8.00 \\
		\end{tabular}
		\caption{Sensitivity of the model to the completeness of the affordance graph.}
		\label{tb:abl_graph}
	\end{center}
\end{table}

\subsection{Details about the HOI Detection Experiment} \label{sec:details_hoid}
In this section we describe the settings that we used for our HOI Detection experiment on HICO-DET and COCO-a (\cref{sec:detection_res}), which differ from the settings of the HOI Recognition experiments (\cref{sec:exp_settings}).

\subsubsection{Experimental Setup} 
The focus of this experiments is Zero-Shot HOI Detection when there are unseen actions. On HICO-DET our training set contains the same unseen actions as the recognition experiment ($\sim$50\% of the total, as described in \cref{sec:exp_settings}), while on COCO-a there are 114 unseen actions, corresponding to 1048 unseen interactions. In both cases there are no unseen object classes, therefore the object branch is removed: since we do not perform zero-shot on objects, we can rely purely on the scores provided by a pre-trained object detector (we use Mask R-CNN \cite{he_mask_2017} with ResNet-50 \cite{he_deep_2016} as backbone). Note that this is possible because we use a model pre-trained on COCO \cite{lin_microsoft_2014}, which has the same object categories as HICO-DET and COCO-a.

\subsubsection{Architectural Changes} 
Contrary to HICO, HICO-DET and COCO-a contain localised information: each interaction in an image refers to a specific person and object, and a bounding box for each is provided. Therefore, we adapted our model to deal with image regions instead of whole images. The object detector provides visual features for every person, every object and every region that represents a possible interaction (i.e., the tightest region that contains both a person and an object). This means that, for each example $ i $, we have three visual feature vectors: $ \boldsymbol{h}_i^{(h)} $, $ \boldsymbol{h}_i^{(o)} $ and $ \boldsymbol{h}_i^{(a)} $ for human, object and action respectively. We compute the interaction representation as
\begin{equation}
\boldsymbol{x}_i = f_{1}([\boldsymbol{h}_i^{(h)}, \boldsymbol{s}_i^{(h)}]) + f_{1}([\boldsymbol{h}_i^{(o)}, \boldsymbol{s}_i^{(o)}]) + f_{1}(\boldsymbol{h}_i^{(a)}) ~,
\end{equation} where $ f_{1} $ is defined as usual as an MLP, $ [\cdot, \cdot] $ indicates concatenation and $ \boldsymbol{s}^{(\cdot)} $ are object classification score vectors returned by the object detector.

\subsubsection{Sampling Interactions}
During training, we keep all detected object bounding boxes and add the ground-truth ones that do not have any match, i.e., there is no detected box whose intersection-over-union (IoU) is greater than 0.5. We keep as positive interaction examples all human-object pairs whose subject and object are correctly classified and overlap with the subject/object (respectively) of a ground-truth interaction (again, the threshold for IoU is 0.5). Among the pairs that are not positive interactions, we sample negative ones, at a rate of 3 negatives per positive (this is a widely used ratio, see for example \cite{peyre_detecting_2019}). At inference time, we only keep human candidates with a confidence score greater than 0.7 and threshold object ones at 0.3. Every possible human-object pair in the image is considered as a candidate interaction and classified by the model.

\subsubsection{Changes to the Training Procedure} When using the regularisation loss $ \mathcal{L}_{REG} $ on HICO-DET, we found it beneficial to only enable it (that is, set $ \rho > 0 $) after the first 5 epochs. This allows the model to learn class representations first, and only later regularise them.

The model is trained with minibatch Stochastic Gradient Descent (SGD) with a learning rate of  $ 0.001 $ and weight decay coefficient of $ 5 \cdot 10^{-4} $. We train our model for a maximum of 10 epochs (due to the high amount of training samples: more than 1.2M interactions, compared to the $\sim$30k training images for HICO) and a batch size of 64, 75\% of which is constituted by negative samples as previously mentioned.

\subsection{Visualisation of Representation Spaces} \label{sec:vis}
In this section we show how the representation spaces vary depending on whether our regularisation loss $ \mathcal{L}_{REG} $ is used (\cref{sec:vis-reg}).

\begin{figure*}
	\centering
	\subfloat[][Unregularised model\label{fig:tsne_base}]{\includegraphics[width=0.9\textwidth]{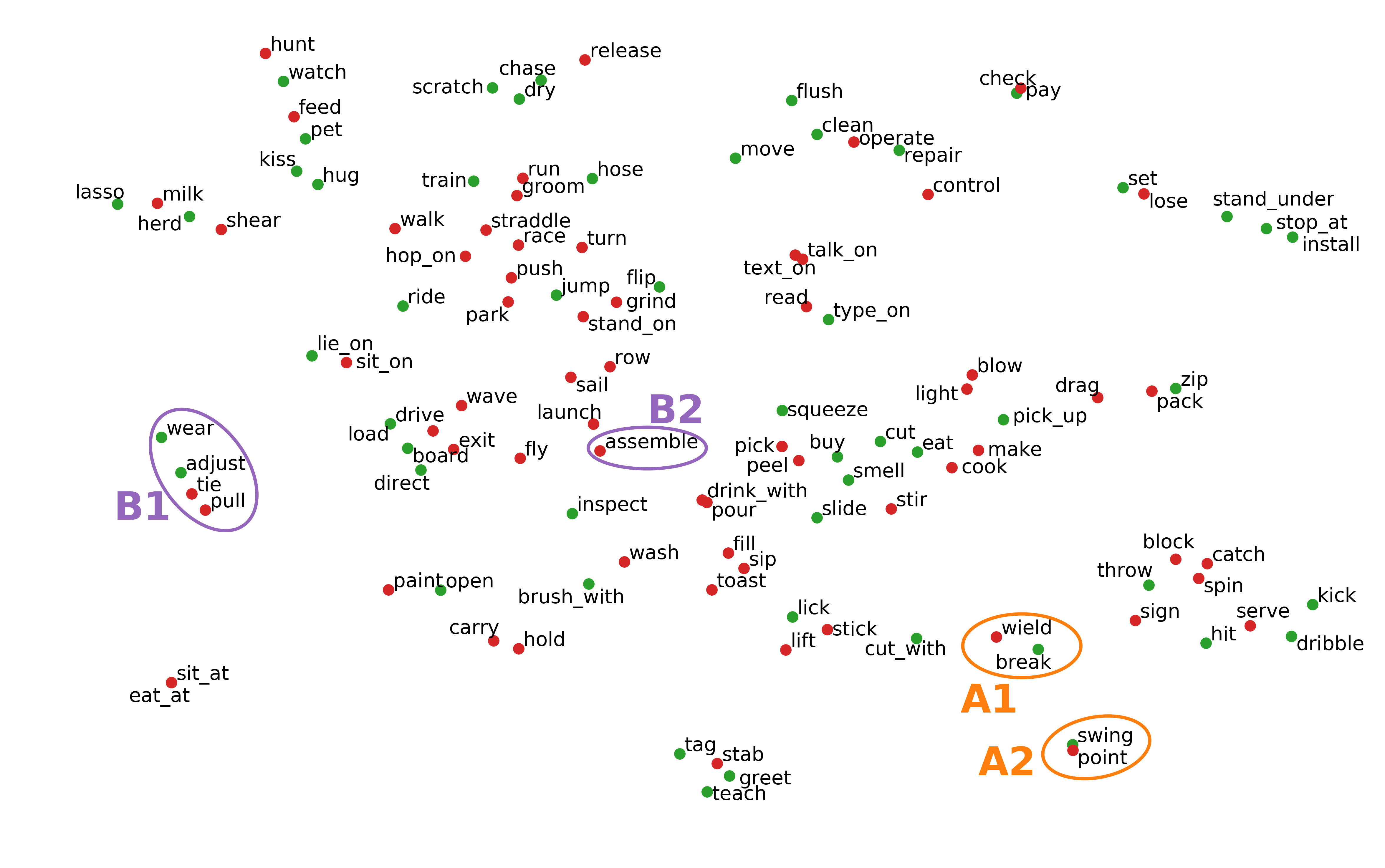}}
	
	\subfloat[][Regularised model\label{fig:tsne_ra}]{\includegraphics[width=0.9\textwidth]{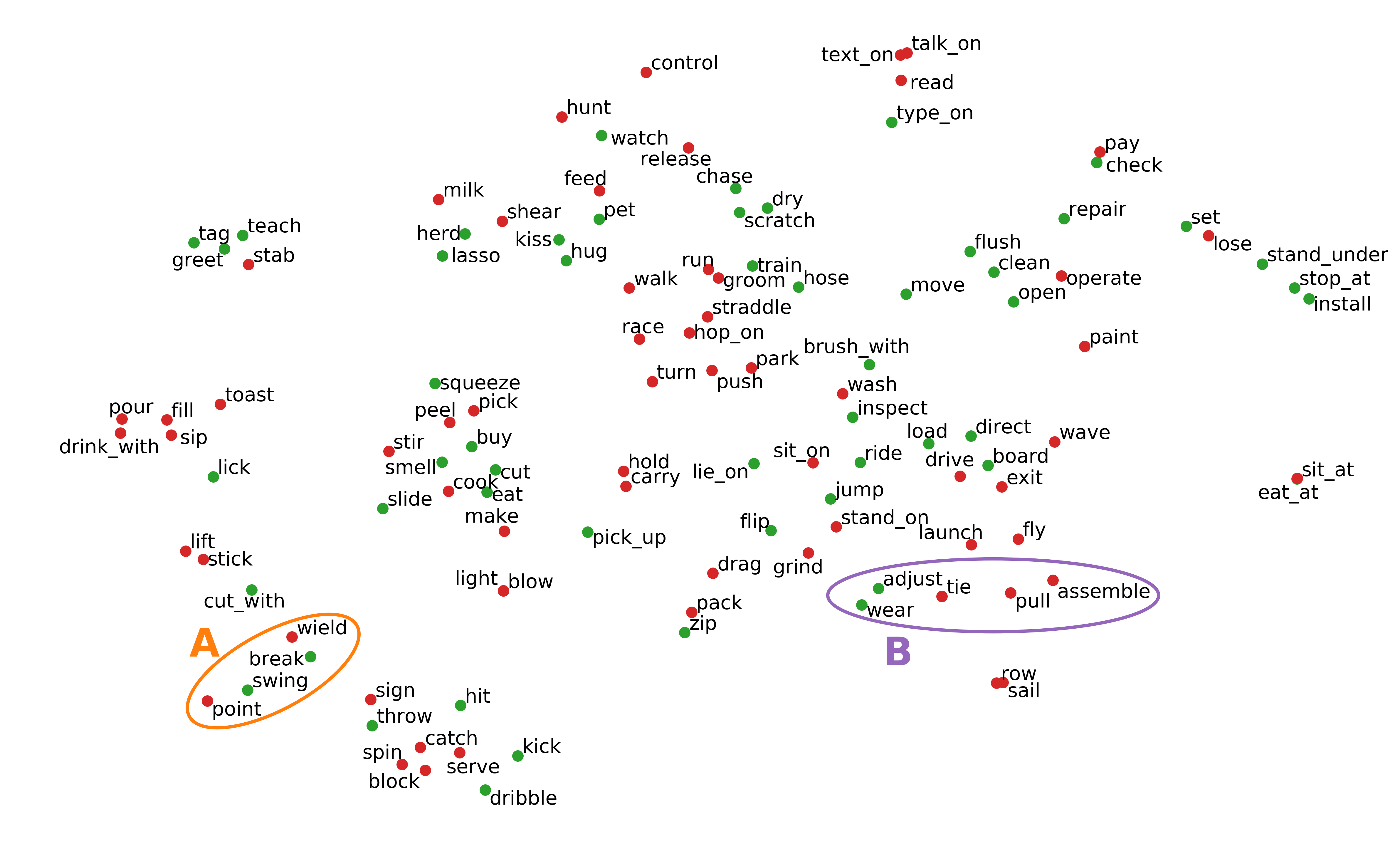}}
	
	\caption{Comparison between unregularised \protect\subref{fig:tsne_base} and regularised \protect\subref{fig:tsne_ra} model. The latter is better at grouping functionally similar actions. Best seen in colour.}
	\label{fig:tsne2}
\end{figure*}

\subsubsection{Effect of Regularisation} \label{sec:vis-reg}

\cref{fig:tsne2} shows action representations obtained with our model on HICO without (\cref{fig:tsne_base}) and with (\cref{fig:tsne_ra}) regularisation. It can be seen that the base model is already quite effective at grouping actions by affordances, as representations are computed through a GCN over the affordance graph. Adding the proposed regularisation further promotes clustering based on functional similarity, i.e., it tends to group together actions based on what objects they can be performed on. For instance, let us consider the group $ wield $, $ break $, $ point $ and $ swing $. Three of them ($ wield $, $ break $ and $ swing $) can be performed on a $ baseball\_bat $, while two ($ swing $ and $ point $) can be performed on a $ remote\_control $. In the unregularised model (\cref{fig:tsne_base}) $ swing $ is correctly clustered with $ point $ (A2), but quite distant from $ wield $ and $ break $ (A1), whereas the regularisation brings the two groups closer to each other, effectively merging them (A). This effect is magnified for $ pull $ and $ assemble $, which can both be performed on a $ kite $. In the base model, $ pull $ is only grouped with actions that can be performed on a $ tie $ (B1), but the regularisation helps in bringing the two clusters (B1 and B2) together because they share a common action (B).

\subsubsection{Are Affordances Captured by Word Embeddings?}

\begin{figure*}[!ht]
	\centering
	\subfloat[][Learnt representation\label{fig:tsne_glove_base}]{\includegraphics[width=.9\textwidth]{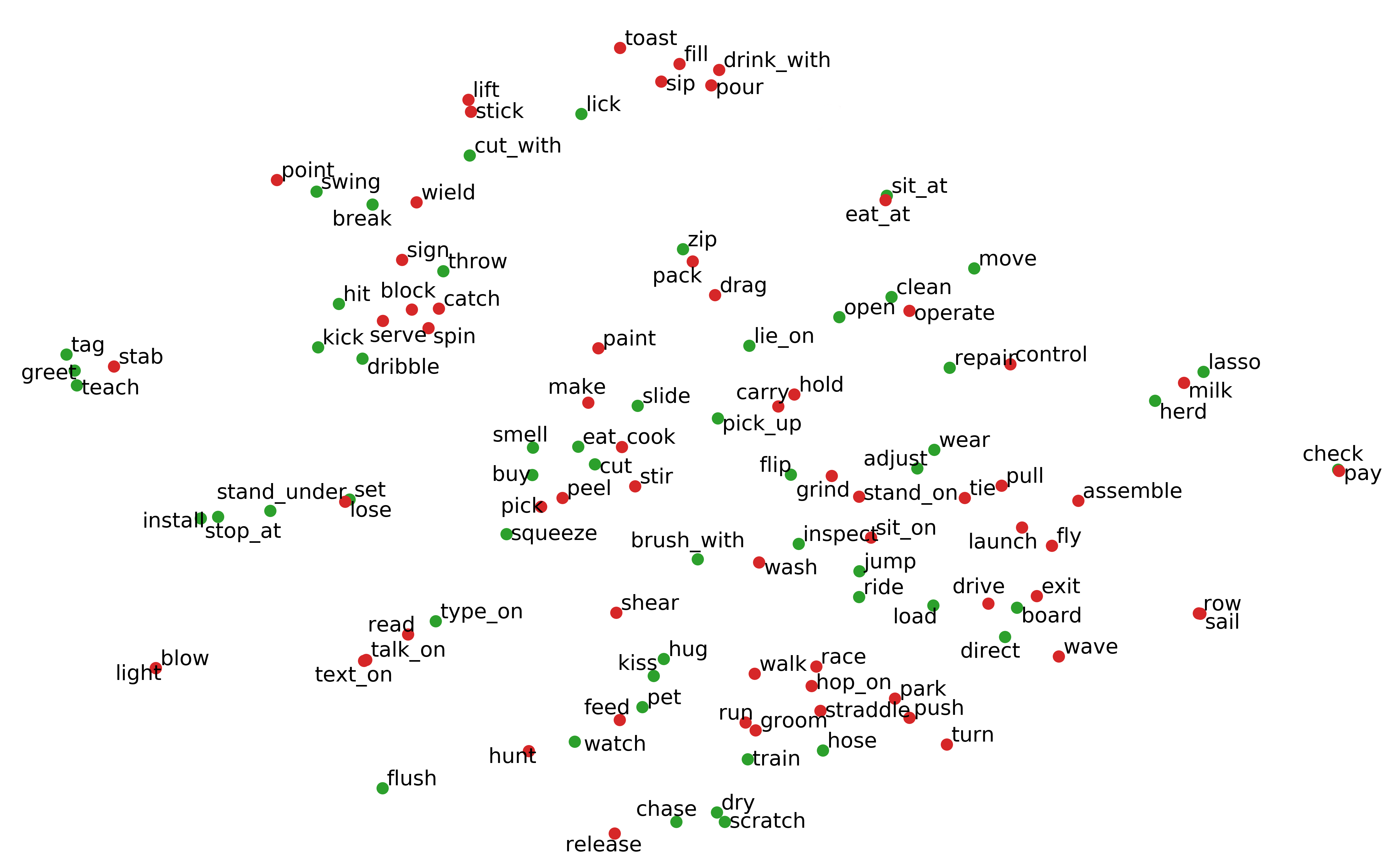}}
	
	\subfloat[][GloVe embedding space\label{fig:tsne_glove_glove}]{\includegraphics[width=.9\textwidth]{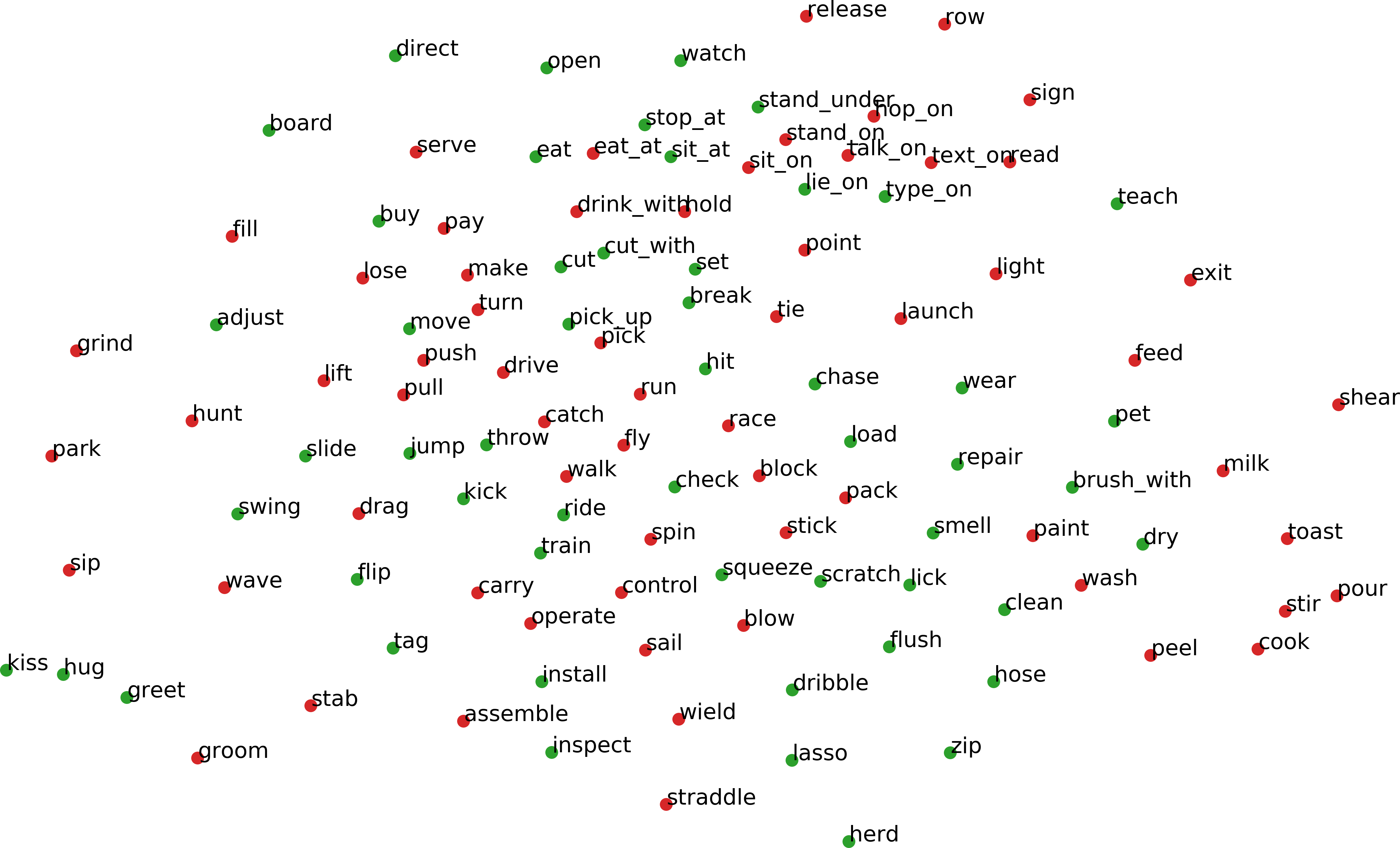}}
	
	\caption{Comparison between the learnt representation space \protect\subref{fig:tsne_glove_base} and the pre-trained word embedding space \protect\subref{fig:tsne_glove_glove}. The proposed approach is effective at grouping actions by affordances, while GloVe embeddings do not capture this relationship. Best seen in colour.}
	\label{fig:tsne_glove}
\end{figure*}

We show a comparison between the learnt representation space and the word embedding space in \cref{fig:tsne_glove}. The figure shows that actions are not clustered by affordance in the word embedding space, further confirming the efficacy of our approach.
\end{document}